# Data-dependent kernels in nearly-linear time

Guy Lever Tom Diethe John Shawe-Taylor

Department of Computer Science
University College London
Gower Street
London WC1E 6BT
UK

October 8, 2018

**Abstract**

We propose a method to efficiently construct data-dependent kernels which can make use of large quantities of (unlabeled) data. Our construction makes an approximation in the standard construction of semi-supervised kernels in Sindhwani et al. (2005). In typical cases these kernels can be computed in nearly-linear time (in the amount of data), improving on the cubic time of the standard construction, enabling large scale semi-supervised learning in a variety of contexts. The methods are validated on semi-supervised and unsupervised problems on data sets containing upto 64'000 sample points.

## 1 Introduction

Semi-supervised methods of inference aim to utilize a large quantity of unlabeled data to assist the learning process. Often this is achieved by using the data to define a data-dependent kernel which captures the geometry of the data distribution, as revealed by the sample. The norm in the reproducing kernel Hilbert space (r.k.h.s.) associated to such a kernel typically includes a data-dependent "intrinsic regularizer" component which captures the smoothness of functions on the data sample. Associated kernel methods such as LapSVM (Belkin et al., 2006) have been shown to achieve state of the art performance in classification.

A drawback of the standard semi-supervised kernel construction, due to Sindhwani et al. (2005), is its large computational cost which is cubic in the number of (unlabeled) data points, rendering the method infeasible for even moderately-sized problems. Several solutions to this problem have been proposed; most apply to particular algorithms only (Zhu and Lafferty, 2005; Collobert et al., 2006; Garcke and Griebel, 2005; Tsang and Kwok, 2006; Sindhwani and Keerthi, 2006; Melacci and Belkin, 2011) or are restricted to the special case of transduction (Mahdaviani et al., 2005). In contrast, we provide efficiently computable data-dependent kernels which can be used in any kernel method.

The kernels we study in this work are obtained by making an approximation in the standard construction of Sindhwani et al. (2005), and can be formed for the same general class of "intrinsic regularizers" considered therein (see the details in Section 2). Our starting point is a given intrinsic regularizer, on functions $h \in \mathbb{R}^{\mathcal{X}}$, of the form $\mathrm{reg}_{\boldsymbol{Q}}(\boldsymbol{h}) := \boldsymbol{h}^\top \boldsymbol{Q}\boldsymbol{h}$, defined on the measurements $\boldsymbol{h} := (h(x_i))_i \in \mathbb{R}^n$ of $h$ at a data sample $\mathcal{X}_\mathcal{S} :=$ $\{x_1, \ldots, x_n\}$, where $\boldsymbol{Q}$ is some symmetric positive semi-definite (often very sparse) matrix. Such regularizers are used in the construction of data-dependent kernels on $\mathcal{X}$ (used, for example, for semi-supervised learning). Implicit in this choice of regularizer is the assumption that the pseudoinverse $\boldsymbol{Q}^+$ is a good covariance on the finite set $\mathcal{X}_{\mathcal{S}}$ (see Theorem 3.1). Our proposed kernels are obtained by replacing the regularizer $\operatorname{reg}_{\boldsymbol{Q}}(\boldsymbol{h})$ with an intrinsic regularizer which measures each function $h$ at a small subsample $\widehat{\mathcal{X}}_{\mathcal{S}} = \{x_{s_1}, \ldots, x_{s_{\widehat{n}}}\} \subset \mathcal{X}_{\mathcal{S}}$ and interpolates the measurement $\widehat{\boldsymbol{h}} := (h(x_{s_i}))_i \in \mathbb{R}^{\widehat{n}}$ to a function $\boldsymbol{h}^* \in \mathbb{R}^n$ over $\mathcal{X}_{\mathcal{S}}$ using the covariance $\boldsymbol{Q}^+$ and uses $\boldsymbol{h}^*$ to approximate $\boldsymbol{h}$: the approximated intrinsic regularizer is thus $\operatorname{reg}_{\boldsymbol{Q}}(\boldsymbol{h}^*)$. For very large $n$ we do not need to measure a function $h$ at all $n$ sample points since $\operatorname{reg}_{\boldsymbol{Q}}(\boldsymbol{h}^*)$ will be a good approximation to $\operatorname{reg}_{\boldsymbol{Q}}(\boldsymbol{h})$ whenever $h$ is in some class of sufficiently smooth functions in the sense specified by $\operatorname{reg}_{\boldsymbol{Q}}$. We then form an r.k.h.s. of functions over the input space, whose norm includes our reduced intrinsic regularizer as a component.

The surprising and useful result we prove is that while the complexity of computing data-dependent kernels is cubic in the number of points at which functions are measured it is only linear in the number of non-zero entries of $\boldsymbol{Q}$, which in typical cases leads to *nearly-linear* complexity in $n$. Thus by disconnecting the number of points used to build the regularization matrix $\boldsymbol{Q}$ (typically all $n$ data points) and the number of points at which functions are measured we are able to practically achieve genuinely large-scale semi-supervised learning. For example when $\boldsymbol{Q}$ is a graph Laplacian our method allows us to use a huge amount of data to build the graph and define the intrinsic regularizer, obtaining a data-dependent kernel on the input space $\mathcal{X}$ in nearly-linear time. This is important since graph building is often not robust at small sample sizes: experimentally we demonstrate a significant advantage can be gained from the ability to exploit a much larger quantity of unlabelled data.

Used with the SVM our kernels can be informally viewed as providing an efficient approximation of LapSVM, and exhibits comparable performance on small datasets. On larger datasets, where LapSVM is infeasible, the method comfortably outperforms the RBF kernel and a more naive implementation of a "budget" LapSVM (defined by discarding the majority of unlabeled data). We also provide an application to clustering.

### 1.1 Preliminaries

We consider the design of kernels suitable for the (semi-)supervised learning problem in which we must infer a regression or classification function $h : \mathcal{X} \to \mathcal{Y}$ mapping instances $x \in \mathcal{X}$ to outputs $y \in \mathcal{Y}$. In particular we suppose there exists a distribution $\mu$ over the set $\mathcal{X} \times \mathcal{Y}$ of labeled instances and that we have a partially labeled sample $\mathcal{S} = \{(x_1, y_1), \ldots, (x_m, y_m)\} \cup \{x_{m+1}, \ldots, x_n\}$ drawn from the product distribution $\mu_{n,m} := \mu^m \times \mu_{\mathcal{X}}^{n-m}$, where $\mu_{\mathcal{X}}$ is the marginal distribution over the instance space $\mathcal{X}$. We denote $\mathcal{X}_{\mathcal{S}} := \{x_i : (x_i, y_i) \in \mathcal{S} \text{ or } x_i \in \mathcal{S}\}$.

For a positive (semi-)definite kernel $K : \mathcal{X} \times \mathcal{X} \to \mathbb{R}$ we denote by $\mathcal{H}_K = \overline{\operatorname{span}\{K(x, \cdot) : x \in \mathcal{X}\}}$ (where completion is w.r.t. the r.k.h.s. norm) its unique r.k.h.s..

For a matrix $\boldsymbol{M}$ we denote by $\boldsymbol{M}^+$ its Moore-Penrose pseudoinverse and by $\operatorname{im}(\boldsymbol{M})$ and $\operatorname{leftnull}(\boldsymbol{M})$ its column and left null spaces. We denote by $\boldsymbol{I}_n$ and $\boldsymbol{1}_n$ the $n \times n$ identity matrix and matrix of all ones respectively, $||\boldsymbol{M}||_\infty = \max_{ij} |M_{ij}|$, and $||\boldsymbol{M}||_2 = \max\{\sqrt{\lambda} : \lambda \text{ is an eigenvalue of } \boldsymbol{M}^\top \boldsymbol{M}\}$ and $\kappa(\boldsymbol{M}) := ||\boldsymbol{M}||_2 ||\boldsymbol{M}^+||_2$. We denote the standard basis in $\mathbb{R}^n$ by $\{\boldsymbol{e}_i\}$. When $\boldsymbol{M}$ is symmetric and positive semi-definite we denote $||\boldsymbol{z}||^2_{\boldsymbol{M}} := \boldsymbol{z}^\top \boldsymbol{M} \boldsymbol{z}$

We view the elements of the set $\mathbb{R}^{\mathcal{V}}$ of real-valued functions on a finite set $\mathcal{V} = \{v_1, \ldots, v_t\}$ as vectors $\boldsymbol{f} \in \mathbb{R}^t$ via $\boldsymbol{f}(v_i) = f_i$.

## 2 Review of semi-supervised kernel methods

We here recall a standard methodology to define a data-dependent kernel for semi-supervised learning in which the norm of the associated r.k.h.s. captures the smoothness of each function w.r.t. the data sample. Given an arbitrary kernel $K : \mathcal{X} \times \mathcal{X} \to \mathbb{R}$, with associated r.k.h.s. $\mathcal{H}_K$, Sindhwani et al. (2005) demonstrate that the space $\mathcal{H}_{\widetilde{K}}$, consisting of functions from $\mathcal{H}_K$, in which the inner product is modified,

$$\langle h, g\rangle_{\widetilde{K}} := \langle h, g\rangle_K + \eta\langle u(h), u(g)\rangle_{\mathcal{U}}, \quad h, g \in \mathcal{H}_K, \tag{1}$$

where $\mathcal{U}$ is any linear space (to be chosen) with positive semi-definite inner product $\langle\cdot,\cdot\rangle_{\mathcal{U}}$, and such that $u : \mathcal{H}_K \to \mathcal{U}$ is a bounded linear map, is an r.k.h.s.. Typically the term $\langle u(h), u(h)\rangle_{\mathcal{U}}$, in the expansion of $||h||^2_{\widetilde{K}}$, acts as a data-dependent "intrinsic regularizer" and captures a notion of smoothness of $h$ over the empirical sample. If we define, $\boldsymbol{h} := (h_i)_i \in \mathbb{R}^n$ as the vector of point evaluations of $h$ on the sample $\mathcal{S}$, $h_i := h(x_i)$, where $x_i \in \mathcal{S}$, then, in particular, when,

$$\langle u(h), u(g)\rangle_{\mathcal{U}} = \boldsymbol{h}^\top \boldsymbol{Q}\boldsymbol{g}, \tag{2}$$

for some symmetric p.s.d. *regularizer matrix* $\boldsymbol{Q}$ then the r.k.h.s. inner product (1) becomes,

$$\langle h, g\rangle_{\widetilde{K}} := \langle h, g\rangle_K + \eta \boldsymbol{h}^\top \boldsymbol{Q}\boldsymbol{g}, \qquad h, g \in \mathcal{H}_K. \tag{3}$$

For such a $\boldsymbol{Q}$, and associated *intrinsic regularization operator* $\mathrm{reg}_{\boldsymbol{Q}} : \boldsymbol{h} \mapsto \boldsymbol{h}^\top\boldsymbol{Q}\boldsymbol{h}$, we say that $h \in \mathcal{H}_K$ is $\mathrm{reg}_{\boldsymbol{Q}}$-smooth whenever $\mathrm{reg}_{\boldsymbol{Q}}(\boldsymbol{h})$ is small. We have,

**Theorem 2.1** (Sindhwani et al. (2005), Proposition 2.2)**.** *The r.k.h.s. $\mathcal{H}_{\widetilde{K}}$ consisting of functions from $\mathcal{H}_K$ with inner product (3) has reproducing kernel $\widetilde{K} : \mathcal{X} \times \mathcal{X} \to \mathbb{R}$ given by,*

$$\widetilde{K}(x, x') = K(x, x') - \eta \boldsymbol{k}_x^\top (\boldsymbol{I} + \eta\boldsymbol{Q}\boldsymbol{K})^{-1}\boldsymbol{Q}\boldsymbol{k}_{x'}, \tag{4}$$

*where $\boldsymbol{k}_x = (K(x_1, x), \ldots, K(x_n, x))^\top$, and $\boldsymbol{K}$ is the $n \times n$ Gram matrix $K_{ij} = K(x_i, x_j)$ for i, $j \leq n$.*

Kernels of the form (4) can be used in any kernel method as a means to achieve the semi-supervised goal of exploiting unlabeled data. One common choice is constructed as follows: given a sample of labeled and unlabeled points $\mathcal{S} = \{(x_1, y_1), \ldots, (x_m, y_m)\} \cup \{x_{m+1}, \ldots, x_n\}$ drawn from the distribution $\mu^m \times \mu_{\mathcal{X}}^{n-m}$ consider the intrinsic regulariser,

$$\begin{aligned}\langle u(h), u(h)\rangle_{\mathcal{U}} &:= \widehat{U}_{\mathcal{S}}(h, h)\\ &= \frac{1}{n(n-1)}\sum_{ij}(h(x_i) - h(x_j))^2 W(x_i, x_j),\end{aligned}$$

where $W : \mathcal{X} \times \mathcal{X} \to \mathbb{R}$ captures similarity or "weight" between data points, for example $W(x, x') = e^{-\gamma||x-x'||^2}$ for some norm $||\cdot||$ over $\mathcal{X}$. Note that $\widehat{U}_{\mathcal{S}}(h, g) = \frac{2}{n(n-1)}\boldsymbol{h}^\top\boldsymbol{L}\boldsymbol{g}$ where $\boldsymbol{L} = \boldsymbol{D} - \boldsymbol{W}$ is a *graph Laplacian* whose *edge weights* are controlled by $\boldsymbol{W} = (W_{ij}) = (W(x_i, x_j))$ and $D_{ij} = \delta_{ij}\sum_k W_{ik}$. This *smoothness functional* is a typical regularizer in semi-supervised learning (Zhu et al., 2003; Belkin et al., 2006, 2004) which punishes functions which do not vary smoothly over the sample. Other choices for the intrinsic regularizer in (3), include many derived from the Laplacian such as the normalised Laplacian $\boldsymbol{L}_{\mathrm{norm}}$ or $\boldsymbol{L}^p$, $\exp(\boldsymbol{L})$, $r(\boldsymbol{L})$ for some real number $p$ or function $r$ (see e.g. Smola and Kondor, 2003)).

The key drawback of the construction (4) (as was pointed out initially for the case of LapSVM in Belkin et al. (2006)) is the $\mathcal{O}(n^3)$ complexity required to invert the matrix $\boldsymbol{I} + \eta\boldsymbol{Q}\boldsymbol{K}$, which renders any derived method such as LapSVM infeasible for even moderately large unlabeled samples. Even once $\boldsymbol{I}+\eta\boldsymbol{Q}\boldsymbol{K}$ is inverted simply evaluating the kernel $\widetilde{K}$ at any pair $(x, x')$ requires an $\mathcal{O}(n^2)$ computation.

## 3 A general method for efficiently constructing data-dependent kernels

Given a partially labeled sample $\mathcal{S}$, we now detail the construction of efficiently computable data-dependent kernels. Recalling the notation of Section 2, we thus suppose that a base kernel $K$ and intrinsic regularization operator $\mathrm{reg}_{\boldsymbol{Q}} : \boldsymbol{h} \mapsto \boldsymbol{h}^\top \boldsymbol{Q}\boldsymbol{h}$ where $\boldsymbol{Q}$ is a p.s.d. regularization matrix, are given (we will consider typical special cases later) and we are interested in constructing an r.k.h.s. $\mathcal{H}_{\breve{K}}$ consisting of functions in $\mathcal{H}_K$ whose inner product achieves the regularization effect of (3) but for which, in contrast to (4), the reproducing kernel $\breve{K}$ is efficiently computable. Approximating (3) subject to computational constraints appears difficult in general and we therefore restrict our attention to a certain specific form of intrinsic inner products which we now describe. We denote a subsample $\widehat{\mathcal{X}}_{\mathcal{S}} = \{x_{s_1}, \ldots, x_{s_{\widehat{n}}}\} \subseteq \mathcal{X}_{\mathcal{S}}$ with $|\widehat{\mathcal{X}}_{\mathcal{S}}| =: \widehat{n} \ll n$ and for a given $h \in \mathcal{H}_K$ denote its evaluation on $\widehat{\mathcal{X}}_{\mathcal{S}}$ by $\widehat{\boldsymbol{h}} := (\widehat{h}_i)_i \in \mathbb{R}^{\widehat{n}}$ where $\widehat{h}_i := h(x_{s_i})$. We then consider those r.k.h.s. $\mathcal{H}_{\breve{K}}$ whose inner products are of the form,

$$\langle h, g\rangle_{\breve{K}} := \langle h, g\rangle_K + \eta \widehat{\boldsymbol{h}}^\top \widehat{\boldsymbol{Q}} \widehat{\boldsymbol{g}} \qquad h, g \in \mathcal{H}_K, \tag{5}$$

where the $\widehat{n} \times \widehat{n}$ symmetric p.s.d. matrix $\widehat{\boldsymbol{Q}}$ is to be chosen. Recalling Theorem 2.1 we see that the kernel $\breve{K}$ is given by,

$$\breve{K}(x, x') = K(x, x') - \eta \widehat{\boldsymbol{k}}_x^\top (\boldsymbol{I}_{\widehat{n}} + \eta \widehat{\boldsymbol{Q}}\widehat{\boldsymbol{K}})^{-1} \widehat{\boldsymbol{Q}} \widehat{\boldsymbol{k}}_{x'}, \tag{6}$$

where, for $x \in \mathcal{X}$, $\widehat{\boldsymbol{k}}_x = (K(x_{s_1}, x), \ldots, K(x_{s_{\widehat{n}}}, x))^\top$, and $\widehat{\boldsymbol{K}}$ is the $\widehat{n} \times \widehat{n}$ Gram matrix $K_{ij} = K(x_{s_i}, x_{s_j})$ for $i, j \le \widehat{n}$. Given $\widehat{\boldsymbol{Q}}$, the complexity of computing (6) is $\mathcal{O}(\widehat{n}^3)$, thus whenever $\widehat{\boldsymbol{Q}}$ is efficiently computable the complexity is substantially less than the $\mathcal{O}(n^3)$ complexity of computing (4). Suppose the subsample $\widehat{\mathcal{X}}_{\mathcal{S}}$ is given[1] and consider the choice of intrinsic regularization matrix $\widehat{\boldsymbol{Q}}$ and associated operator $\mathrm{reg}_{\widehat{\boldsymbol{Q}}} : \widehat{\boldsymbol{h}} \mapsto \widehat{\boldsymbol{h}}^\top \widehat{\boldsymbol{Q}} \widehat{\boldsymbol{h}}$. The most straightforward form of this approach would be, given $\widehat{\mathcal{X}}_{\mathcal{S}}$, to discard all remaining data instances $\mathcal{X}_{\mathcal{S}} \backslash \widehat{\mathcal{X}}_{\mathcal{S}}$ and construct $\widehat{\boldsymbol{Q}}$ using only the subsample $\widehat{\mathcal{X}}_{\mathcal{S}}$ – typically, for example, $\widehat{\boldsymbol{Q}}$ might be derived from the Laplacian of a graph built on the subset $\widehat{\mathcal{X}}_{\mathcal{S}}$. In discarding almost all unlabeled data no advantage can be gained from it and this simplistic method should act as a benchmark which any proposed method should improve upon. The task is to choose an $\widehat{n} \times \widehat{n}$ matrix $\widehat{\boldsymbol{Q}}$ which achieves the effect of (4) exploiting all unlabeled data. It is perhaps surprising that such a $\widehat{\boldsymbol{Q}}$ exists: for example, when $\boldsymbol{Q}$ is an $n \times n$ Laplacian of a graph $\mathcal{G}$ constructed on all of $\mathcal{X}_{\mathcal{S}}$, we can find an $\widehat{n} \times \widehat{n}$ regularization matrix $\widehat{\boldsymbol{Q}}$ whose associated regularization operator involves the full structure of the graph $\mathcal{G}$.

To motivate a natural choice for $\widehat{\boldsymbol{Q}}$ in (5) we first recall some well-known facts regarding the duality between positive semi-definite regularization operators on spaces of functions and kernels on their domain defined by their Green's functions (e.g. Smola et al., 1998). The following is a special case for finite input sets (the proof is given in the Appendix):

**Theorem 3.1.** *(e.g. Smola and Kondor, 2003, Theorem 4) Given a finite set of points $\mathcal{V} = \{v_1, \ldots, v_t\}$, consider $\boldsymbol{h} \in \mathbb{R}^{\mathcal{V}}$ as a vector $\boldsymbol{h} \in \mathbb{R}^t$ via $\boldsymbol{h}(v_i) := \boldsymbol{h}^\top \boldsymbol{e}_i = h_i$. Consider further a regularization operator on such functions, $\mathrm{reg} : \mathbb{R}^t \to \mathbb{R}$ given by $\mathrm{reg}(\boldsymbol{h}) = \boldsymbol{h}^\top \boldsymbol{R}\boldsymbol{h}$, where $\boldsymbol{R}$ is a symmetric positive semi-definite matrix. Then the Hilbert space $\mathcal{H} = \mathrm{im}(\boldsymbol{R}) \subseteq \mathbb{R}^t$ of real-valued functions on $\mathcal{V}$ with inner product $\langle \boldsymbol{h}, \boldsymbol{g}\rangle_{\mathcal{H}} = \boldsymbol{h}^\top \boldsymbol{R}\boldsymbol{g}$ is an r.k.h.s. whose reproducing kernel $K : \mathcal{V} \times \mathcal{V} \to \mathbb{R}$ is given by $\boldsymbol{R}^+$, i.e. such that $K(v_i, v_j) := R^+_{ij} = \boldsymbol{e}_i^\top \boldsymbol{R}^+ \boldsymbol{e}_j$.*

The Green's function in this case simply being the matrix pseudoinverse of the regularization operator $\boldsymbol{R}$. Thus sensible regularization operators on functions over finite sets define sensible reproducing kernels via their pseudoinverse. A natural choice for the regularization operator $\widehat{\boldsymbol{Q}}$ is immediately motivated by the above observations:

[1] A random subsample of $\mathcal{X}_{\mathcal{S}}$ seems sensible as it would ensure that $\widehat{\mathcal{X}}_{\mathcal{S}}$ is an i.i.d. distribution from the underlying data-generating distribution.

for the given intrinsic regularizer $\boldsymbol{Q}$ we can view $\boldsymbol{Q}^+$ as a kernel on $\mathcal{X}_\mathcal{S}$ via $\boldsymbol{Q}^+(x_i, x_j) = Q^+_{ij}$. In particular, the submatrix $\boldsymbol{Q}^+|_{\widehat{\mathcal{X}}_\mathcal{S}} = (Q^+_{s_i,s_j} \ : \ i, j \le \widehat{n})$, being the gram matrix of the restriction of this kernel to $\widehat{\mathcal{X}}_\mathcal{S}$, is always a valid positive semi-definite kernel on $\widehat{\mathcal{X}}_\mathcal{S}$ which captures precisely the affinities on $\widehat{\mathcal{X}}_\mathcal{S}$ induced by $\boldsymbol{Q}^+$. Thus, recalling the duality of Theorem 3.1, our proposed choice of regularizer $\widehat{\boldsymbol{Q}}$ is,

$$\widehat{\boldsymbol{Q}} = \left(\boldsymbol{Q}^+|_{\widehat{\mathcal{X}}_\mathcal{S}}\right)^+, \tag{7}$$

i.e. $\widehat{Q}^+_{ij} = Q^+_{s_i,s_j}$, and its associated operator $\operatorname{reg}_{\widehat{\boldsymbol{Q}}} : \widehat{\boldsymbol{h}} \mapsto \widehat{\boldsymbol{h}}^\top \widehat{\boldsymbol{Q}} \widehat{\boldsymbol{h}}$, is a natural regularizer for functions on $\widehat{\mathcal{X}}_\mathcal{S}$. Since $\widehat{\boldsymbol{Q}}^+$ given by (7) is the submatrix of the pseudoinverse of the $n \times n$ matrix $\boldsymbol{Q}$ it might seem that $\widehat{\boldsymbol{Q}}$ is not efficiently computable but in Section 4 we will see that for typical choices of $\boldsymbol{Q}$, an $\epsilon$-approximation to the kernel $\widehat{\boldsymbol{Q}}^+$ is computable in nearly-linear time.

### 3.1 Interpreting the intrinsic regularizer

So far we have motivated our choice of regularization matrix $\widehat{\boldsymbol{Q}} = \left(\boldsymbol{Q}^+|_{\widehat{\mathcal{X}}_\mathcal{S}}\right)^+$ in (6) by demonstrating that its pseudoinverse $\widehat{\boldsymbol{Q}}^+$ is a natural kernel on the subset $\widehat{\mathcal{X}}_\mathcal{S}$, and invoking Theorem 3.1. We now give the key result interpreting the intrinsic inner product in terms of the interpolation of functions defined on $\widehat{\mathcal{X}}_\mathcal{S}$ to $\mathcal{X}_\mathcal{S}$.

We first introduce some notation: note that we can reorder the set $\mathcal{X}_\mathcal{S} = \{x_1, \ldots, x_n\}$ such that w.l.o.g. we can consider $\widehat{\mathcal{X}}_\mathcal{S} = \{x_1, \ldots, x_{\widehat{n}}\}$. We then write $\overline{\mathcal{X}}_\mathcal{S} = \mathcal{X}_\mathcal{S} \backslash \widehat{\mathcal{X}}_\mathcal{S}$, $\bar{n} := |\overline{\mathcal{X}}_\mathcal{S}| = n - \widehat{n}$ and $\boldsymbol{Q} = \begin{pmatrix} \boldsymbol{Q}_{\widehat{n}\widehat{n}} & \boldsymbol{Q}_{\widehat{n}\bar{n}} \\ \boldsymbol{Q}_{\bar{n}\widehat{n}} & \boldsymbol{Q}_{\bar{n}\bar{n}} \end{pmatrix}$ where $\boldsymbol{Q}_{\widehat{n}\bar{n}} = (Q_{ij} \ : \ x_i \in \widehat{\mathcal{X}}_\mathcal{S}, x_j \in \overline{\mathcal{X}}_\mathcal{S})$ and $\boldsymbol{Q}_{\widehat{n}\widehat{n}}$, $\boldsymbol{Q}_{\bar{n}\bar{n}}$ etc. are defined analogously. For a function $h \in \mathbb{R}^\mathcal{X}$ define $\operatorname{int}_{\boldsymbol{Q}}(\widehat{\boldsymbol{h}}) := \operatorname{argmin}_{\boldsymbol{f} \in \mathbb{R}^{\mathcal{X}_\mathcal{S}}} \{\operatorname{reg}_{\boldsymbol{Q}}(\boldsymbol{f}) \ : \ \boldsymbol{f}|_{\widehat{\mathcal{X}}_\mathcal{S}} = \widehat{\boldsymbol{h}}\}$ the minimum (semi-)norm interpolants of $\widehat{\boldsymbol{h}}$. We have:

**Theorem 3.2.** *Suppose $\boldsymbol{Q}$ is such that the generalized Schur complement $\boldsymbol{Q}_{\widehat{n}\widehat{n}} - \boldsymbol{Q}_{\widehat{n}\bar{n}} \boldsymbol{Q}^+_{\bar{n}\bar{n}} \boldsymbol{Q}_{\bar{n}\widehat{n}}$ is nonsingular*[2] *then, for any $h, g \in \mathbb{R}^\mathcal{X}$, the intrinsic inner product in (5) with $\widehat{\boldsymbol{Q}}$ defined by (7) satisfies,*

$$\widehat{\boldsymbol{h}}^\top \widehat{\boldsymbol{Q}} \widehat{\boldsymbol{g}} = (\boldsymbol{h}^*)^\top \boldsymbol{Q} \boldsymbol{g}^*,$$

*where $\boldsymbol{h}^*$ and $\boldsymbol{g}^*$ are any elements of the sets of interpolants $\operatorname{int}_{\boldsymbol{Q}}(\widehat{\boldsymbol{h}})$ and $\operatorname{int}_{\boldsymbol{Q}}(\widehat{\boldsymbol{g}})$.*

*Proof.* Consider some $h, g \in \mathbb{R}^\mathcal{X}$ so that $\boldsymbol{h}^* \in \operatorname{int}_{\boldsymbol{Q}}(\widehat{\boldsymbol{h}})$, $\boldsymbol{g}^* \in \operatorname{int}_{\boldsymbol{Q}}(\widehat{\boldsymbol{g}})$. For any $\boldsymbol{f} \in \mathbb{R}^{\mathcal{X}_\mathcal{S}}$ note that,

$$\operatorname{reg}_{\boldsymbol{Q}}(\boldsymbol{f}) = \begin{pmatrix} \boldsymbol{f}|_{\widehat{\mathcal{X}}_\mathcal{S}} \\ \boldsymbol{f}|_{\overline{\mathcal{X}}_\mathcal{S}} \end{pmatrix}^\top \begin{pmatrix} \boldsymbol{Q}_{\widehat{n}\widehat{n}} & \boldsymbol{Q}_{\widehat{n}\bar{n}} \\ \boldsymbol{Q}_{\bar{n}\widehat{n}} & \boldsymbol{Q}_{\bar{n}\bar{n}} \end{pmatrix} \begin{pmatrix} \boldsymbol{f}|_{\widehat{\mathcal{X}}_\mathcal{S}} \\ \boldsymbol{f}|_{\overline{\mathcal{X}}_\mathcal{S}} \end{pmatrix}, \tag{8}$$

and differentiating (8) w.r.t. $\boldsymbol{f}|_{\overline{\mathcal{X}}_\mathcal{S}}$, setting this to zero when $\boldsymbol{f} = \boldsymbol{h}^*$ and setting $\boldsymbol{h}^*|_{\widehat{\mathcal{X}}_\mathcal{S}} = \widehat{\boldsymbol{h}}$ we obtain,

$$\boldsymbol{Q}_{\bar{n}\bar{n}} \boldsymbol{h}^*|_{\overline{\mathcal{X}}_\mathcal{S}} = -\boldsymbol{Q}_{\bar{n}\widehat{n}} \widehat{\boldsymbol{h}} \tag{9}$$

so that $\boldsymbol{h}^* = \begin{pmatrix} \widehat{\boldsymbol{h}} \\ -\boldsymbol{Q}^+_{\bar{n}\bar{n}} \boldsymbol{Q}_{\bar{n}\widehat{n}} \widehat{\boldsymbol{h}} + \boldsymbol{u} \end{pmatrix}$, for some $\boldsymbol{u} \in \operatorname{leftnull}(\boldsymbol{Q}_{\bar{n}\bar{n}})$, and we can similarly derive $\boldsymbol{g}^* = \begin{pmatrix} \widehat{\boldsymbol{g}} \\ -\boldsymbol{Q}^+_{\bar{n}\bar{n}} \boldsymbol{Q}_{\bar{n}\widehat{n}} \widehat{\boldsymbol{g}} + \boldsymbol{v} \end{pmatrix}$,

[2] This is guaranteed, for example, when $\boldsymbol{Q}$ is positive definite.

for some $\boldsymbol{v} \in \operatorname{leftnull}(\boldsymbol{Q}_{\bar{n}\bar{n}})$. We obtain,

$$\begin{aligned}\boldsymbol{h}^{*\top}\boldsymbol{Q}\boldsymbol{g}^* &= \widehat{\boldsymbol{h}}^\top \left(\boldsymbol{Q}_{\widehat{n}\widehat{n}} - \boldsymbol{Q}_{\widehat{n}\bar{n}}\boldsymbol{Q}_{\bar{n}\bar{n}}^+\boldsymbol{Q}_{\bar{n}\widehat{n}}\right)\widehat{\boldsymbol{g}} \\ &\quad + \boldsymbol{u}^\top\boldsymbol{Q}_{\bar{n}\widehat{n}}\widehat{\boldsymbol{h}} + \boldsymbol{v}^\top\boldsymbol{Q}_{\bar{n}\widehat{n}}\widehat{\boldsymbol{g}} \\ &= \widehat{\boldsymbol{h}}^\top\left(\boldsymbol{Q}^+|_{\widehat{\mathcal{X}}_\mathcal{S}}\right)^+\widehat{\boldsymbol{g}} = \widehat{\boldsymbol{h}}^\top\widehat{\boldsymbol{Q}}\widehat{\boldsymbol{g}}.\end{aligned}$$

The final line following from the formula for generalized inverses of partitioned matrices[3] (see e.g. Rohde, 1965) and since (9) implies that $\boldsymbol{Q}_{\bar{n}\widehat{n}}\widehat{\boldsymbol{h}} \perp \operatorname{leftnull}(\boldsymbol{Q}_{\bar{n}\bar{n}}) \ni \boldsymbol{u}$ (and we similarly remove the term in $\boldsymbol{v}$). □

In particular, the smoothness that $\operatorname{reg}_{\widehat{\boldsymbol{Q}}}(\widehat{\boldsymbol{h}})$ measures is therefore the $\operatorname{reg}_{\boldsymbol{Q}}$-smoothness of any minimum (semi-)norm interpolant $\boldsymbol{h}^*$ of $\widehat{\boldsymbol{h}}$. There is also a Bayesian interpretation: the $\operatorname{reg}_{\widehat{\boldsymbol{Q}}}$-smoothness of $\widehat{\boldsymbol{h}}$ is the $\operatorname{reg}_{\boldsymbol{Q}}$-smoothness of the posterior mean of Bayesian inference in the GP using covariance $\boldsymbol{Q}^+$ with observations $\widehat{\boldsymbol{h}}$ sampled at $\widehat{\mathcal{X}}_\mathcal{S}$ in the limit of no noise – there is a well-known equivalence with the minimum semi-norm interpolant.

### 3.1.1 Spcialization to graph Laplacian-based regularizers

Via Theorem 3.2 we see that $\operatorname{reg}_{\widehat{\boldsymbol{Q}}}$ takes into account the whole of the data sample (whenever $\boldsymbol{Q}$ does). We now expand upon this in the common case when $\boldsymbol{Q}$ is (derived from) a graph Laplacian. Given a graph $\mathcal{G} = (\mathcal{V}, \mathcal{E})$ constructed on $\mathcal{X}_\mathcal{S}$ (i.e. there is a bijection $\mathcal{X}_\mathcal{S} \to \mathcal{V}$), suppose that $\operatorname{reg}_{\boldsymbol{Q}}$ measures smoothness of functions over the vertices $\mathcal{V}$ w.r.t. the graph structure (as explained in Section 2), the typical example being when $\boldsymbol{Q}$ is a Laplacian. The $\widehat{\boldsymbol{Q}}$-smoothness $\operatorname{reg}_{\widehat{\boldsymbol{Q}}}(\widehat{\boldsymbol{h}}) = \widehat{\boldsymbol{h}}^\top\widehat{\boldsymbol{Q}}\widehat{\boldsymbol{h}}$ of $\widehat{\boldsymbol{h}} \in \mathbb{R}^{\widehat{\mathcal{X}}_\mathcal{S}}$ is then small whenever $\widehat{\boldsymbol{h}}$ admits an extension to the full vertex set $\mathcal{V}$ which respects the structure of $\mathcal{G}$ as illustrated in Figure 1.

# 4 Complexity analysis

We now show that for typical choices of intrinsic regularization matrices $\boldsymbol{Q}$ (an approximation to) our kernel $\breve{K}$ is efficiently computable. If $\widehat{\boldsymbol{Q}}$ is computed then there is a one time $\mathcal{O}(\widehat{n}^3)$ cost to construct $(\boldsymbol{I}_{\widehat{n}} + \eta\widehat{\boldsymbol{Q}}\widehat{\boldsymbol{K}})^{-1}$ following which kernel evaluations can be computed in $\mathcal{O}(\widehat{n}^2)$ time. Therefore it is required to demonstrate the complexity of computing $\widehat{\boldsymbol{Q}}$.

We first consider the case when $\boldsymbol{Q}$ is symmetric, diagonally dominant sparse matrix with $s$ non-zero entries, and suppose for simplicity that $s \geq n$. In this case we show that there is an algorithm with complexity $\mathcal{O}(\widehat{n}s\log n(\log\log n)^2\log\frac{1}{\epsilon} + \widehat{n}^2 n)$ which returns an $\epsilon$-approximation $\boldsymbol{A}$ to the kernel matrix $\widehat{\boldsymbol{Q}}^+$. We need the following lemma which is a recent example of nearly-linear time solvers for sparse symmetric diagonally dominant linear systems pioneered by Spielman and Teng (2006).

**Lemma 4.1.** *(Koutis et al., 2011)[4] Given a symmetric diagonally dominant $n \times n$ matrix $\boldsymbol{M}$ with $s$ non-zero entries and a vector $\boldsymbol{b} \in \mathbb{R}^n$ there exists an algorithm which in expected time $\mathcal{O}(s\log n(\log\log n)^2\log\frac{1}{\epsilon})$ computes $\boldsymbol{z} \in \mathbb{R}^n$ satisfying $||\boldsymbol{z} - \boldsymbol{M}^+\boldsymbol{b}||_{\boldsymbol{M}} < \epsilon||\boldsymbol{M}^+\boldsymbol{b}||_{\boldsymbol{M}}$.*

We can now prove the following:

[3] Which, when $\boldsymbol{Q}$ is positive definite reduces to the well-known formula.

[4] Published papers with similar guarantees are (Koutis et al., 2010; Spielman and Teng, 2006).

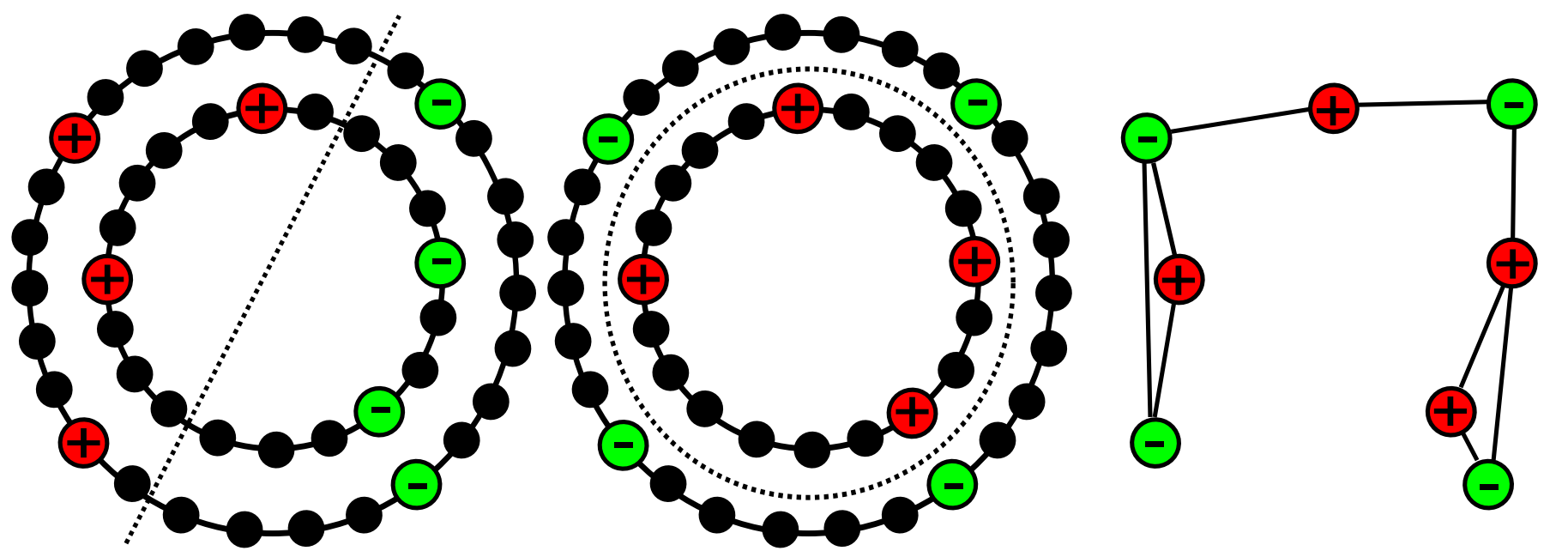

(a) non-$\widehat{\boldsymbol{Q}}$-smooth labeling (b) $\widehat{\boldsymbol{Q}}$-smooth labeling (c) Graph on reduced sample

Figure 1: Artificial illustration: concentric circles. A 2-NN graph built on the large data sample (black spots connected by edges) captures the underlying structure of the two concentric circles defining the two classes. Suppose $\boldsymbol{Q}$ captures smoothness on this graph. The subsample $\widehat{\mathcal{X}}_{\mathcal{S}}$ is highlighted red and green. The hypothesis of Figure 1(a) (the separating hyperplane is shown by the dotted line) is non-$\widehat{\boldsymbol{Q}}$-smooth: there is no $\boldsymbol{Q}$-smooth extension of the labeling of $\widehat{\mathcal{X}}_{\mathcal{S}}$ to the full graph. The hypothesis of Figure 1(b), separating the two classes, is $\widehat{\boldsymbol{Q}}$-smooth: there exists a $\boldsymbol{Q}$-smooth extension of the labeling of $\widehat{\mathcal{X}}_{\mathcal{S}}$ to the full graph. Figure 1(c): a 2-NN graph built on $\widehat{\mathcal{X}}_{\mathcal{S}}$ does not capture the structure of the data-distribution and the correct labeling is not smooth w.r.t. this graph.

**Theorem 4.2.** *Given a symmetric diagonally dominant $n \times n$ matrix $\boldsymbol{Q}$ with $s$ non-zero entries an approximation $\boldsymbol{A}$ to the kernel matrix $\widehat{\boldsymbol{Q}}^+$ on $\widehat{\mathcal{X}}_{\mathcal{S}}$ can be computed in expected time $\mathcal{O}(\widehat{n}s\log n(\log\log n)^2\log\frac{1}{\epsilon}+\widehat{n}^2 n)$ where,*

$$|A_{ij}-\widehat{Q}^+_{ij}|<\epsilon Q^+_{s_is_i}Q^+_{s_js_j},$$

*and thus in sup norm,*

$$||\boldsymbol{A}-\widehat{\boldsymbol{Q}}^+||_\infty<\epsilon\sup_{\{i\,:\,x_i\in\widehat{\mathcal{X}}_{\mathcal{S}}\}}(Q^+_{ii})^2, \tag{10}$$

*and in spectral norm,*

$$||\boldsymbol{A}-\widehat{\boldsymbol{Q}}^+||_2<\epsilon\sum_{\{i\,:\,x_i\in\widehat{\mathcal{X}}_{\mathcal{S}}\}}(Q^+_{ii})^2. \tag{11}$$

*Proof.* We begin by making $\widehat{n}$ calls to the solver of Koutis et al. (2011) to solve the equations

$$\boldsymbol{Q}\boldsymbol{z}_i=\boldsymbol{e}_{s_i} \tag{12}$$

for each $i$ where $\boldsymbol{x}_{s_i}\in\widehat{\mathcal{X}}_{\mathcal{S}}$, giving $\boldsymbol{z}_i$ such that $||\boldsymbol{z}_i-\boldsymbol{Q}^+\boldsymbol{e}_{s_i}||_{\boldsymbol{Q}}\le\epsilon||\boldsymbol{Q}^+\boldsymbol{e}_{s_i}||_{\boldsymbol{Q}}=\epsilon Q^+_{s_is_i}$ in total time $\mathcal{O}(\widehat{n}s\log n(\log\log n)^2\log\frac{1}{\epsilon})$ by Lemma 4.1. Now let $\boldsymbol{Z}:=\begin{pmatrix}\boldsymbol{z}_1 & \dots & \boldsymbol{z}_{\widehat{n}}\end{pmatrix}$ and

$$\boldsymbol{A}:=\boldsymbol{Z}^\top\boldsymbol{Q}\boldsymbol{Z},$$

and note that $\boldsymbol{A}$ can be computed with $\mathcal{O}(s\widehat{n}+\widehat{n}^2n)$ operations since $\boldsymbol{Q}$ has $s$ non-zero entries. Now note that $|\widehat{Q}^+_{ij}-A_{ij}|=|Q^+_{s_is_j}-A_{ij}|=|\boldsymbol{e}_{s_i}^\top\boldsymbol{Q}^+\boldsymbol{e}_{s_j}-\boldsymbol{z}_i^\top\boldsymbol{Q}\boldsymbol{z}_j|=|(\boldsymbol{Q}^+\boldsymbol{e}_{s_i}-\boldsymbol{z}_i)^\top\boldsymbol{Q}\boldsymbol{Q}^+\boldsymbol{e}_{s_j}+(\boldsymbol{Q}^+\boldsymbol{e}_{s_j}-\boldsymbol{z}_j)^\top\boldsymbol{Q}\boldsymbol{Q}^+\boldsymbol{e}_{s_i}-$

$(\boldsymbol{Q}^{+}\boldsymbol{e}_{s_i} - \boldsymbol{z}_i)^\top \boldsymbol{Q}(\boldsymbol{Q}^{+}\boldsymbol{e}_{s_j} - \boldsymbol{z}_j)| \leq ||\boldsymbol{Q}^{+}\boldsymbol{e}_{s_i} - \boldsymbol{z}_i||_{\boldsymbol{Q}} Q^{+}_{s_j s_j} + ||\boldsymbol{Q}^{+}\boldsymbol{e}_{s_j} - \boldsymbol{z}_j||_{\boldsymbol{Q}} Q^{+}_{s_i s_i} + ||\boldsymbol{Q}^{+}\boldsymbol{e}_{s_i} - \boldsymbol{z}_i||_{\boldsymbol{Q}} ||\boldsymbol{Q}^{+}\boldsymbol{e}_{s_j} - \boldsymbol{z}_j||_{\boldsymbol{Q}} < (2\epsilon + \epsilon^2) Q^{+}_{s_i s_i} Q^{+}_{s_j s_j}$ which (after rescaling $\epsilon' = 2\epsilon + \epsilon^2$) proves (10). Now note that,

$$\begin{aligned}
||\boldsymbol{A} - \widehat{\boldsymbol{Q}}^{+}||_2 &= \sup_{\{\boldsymbol{x}\,:\,||\boldsymbol{x}||\leq 1\}} \left| \boldsymbol{x}^\top (\boldsymbol{A} - \widehat{\boldsymbol{Q}}^{+}) \boldsymbol{x} \right| \\
&= \sup_{\{\boldsymbol{x}\,:\,||\boldsymbol{x}||\leq 1\}} \left| \sum_{ij \leq \widehat{n}} x_i x_j (A_{ij} - \widehat{Q}^{+}_{ij}) \right| \\
&\leq \epsilon \sup_{\{\boldsymbol{x}\,:\,||\boldsymbol{x}||\leq 1\}} \sum_{i \leq \widehat{n}} |x_i Q^{+}_{s_i s_i}| \sum_{j \leq \widehat{n}} |x_j Q^{+}_{s_j s_j}| \\
&\leq \epsilon \sum_{i \leq \widehat{n}} (Q^{+}_{s_i s_i})^2,
\end{aligned}$$

which proves (11). □

The linear solvers used to compute the regularizer $\widehat{\boldsymbol{Q}}$, utilise low stretch spanning tree preconditioners. In practice we use a recent practical implementation (Koutis, 2011) of these ideas (though not precisely the algorithm attaining the guarantee above) and achieve linear-time scaling in practice. We can also derive a similar result for an approximation of the kernel $\breve{K}$, and we essentially incur an additional logarithmic dependence upon $\frac{1}{\lambda_{\min}}$ where $\lambda_{\min} := \min\{\lambda\ :\ \lambda \text{ is an eigenvalue of } \widehat{\boldsymbol{Q}}^{-1} + \eta\widehat{\boldsymbol{K}}\}$. The following theorem is proved in the appendix:

**Theorem 4.3.** *Given a symmetric diagonally dominant, $n \times n$ matrix $\boldsymbol{Q}$ with $s$ non-zero entries let $\widehat{\boldsymbol{Q}}$ be as defined in (7) and suppose further that $\widehat{\boldsymbol{Q}}$ is positive definite. Let $\lambda_{\min} := \min\{\lambda\ :\ \lambda$ is an eigenvalue of $\widehat{\boldsymbol{Q}}^{-1} + \eta\widehat{\boldsymbol{K}}\}$. If $\epsilon \ll \lambda_{\min}$, then an approximation $\breve{K}_{\boldsymbol{A}}$ to the kernel $\breve{K}$ defined by (6), and where $K$ satisfies $\sup_{x\in\mathcal{X}} K(x,x) = \kappa < \infty$, can be computed in expected time $\mathcal{O}(\widehat{n}s \log n (\log\log n)^2 \log \frac{q\widehat{n}\eta\kappa}{\epsilon\lambda_{\min}} + \widehat{n}^2 n)$, where $q := \sum_{\{i\,:\,x_i \in \widehat{\mathcal{X}}_S\}} (Q^{+}_{ii})^2$, such that*

$$\sup_{x,x' \in \mathcal{X}} |\breve{K}(x,x') - \breve{K}_{\boldsymbol{A}}(x,x')| < \epsilon + h.o.t.,$$

*where h.o.t. denotes smaller terms in $\epsilon^2$ or greater.*

## 4.1 Laplacians, higher order regularizers and amplified resistances

In the case when $\boldsymbol{Q} = \boldsymbol{L}$, a sparse (connected) graph Laplacian, as is typically the case in semi-supervised learning applications, Theorem 4.2 demonstrates that we can approximate the kernel $\widehat{\boldsymbol{Q}}^{+}$ well. By applying simple transforms to the linear systems solved in the proof of Theorem 4.2, very similar results will hold for the normalized Laplacian $\boldsymbol{L}_{\text{norm}} = \boldsymbol{D}^{-1/2}\boldsymbol{L}\boldsymbol{D}^{-1/2}$ or other regualrizers obtained from simple transforms. Recent theoretical and practical results (Nadler et al., 2009; Zhou and Belkin, 2011; von Luxburg et al., 2010) demonstrate some problems with using $\boldsymbol{L}$ as a regularizer: for example, the solution of Laplacian regularised empirical risk minimization degenerates to a constant function with spikes at labeled points in the limit of large data whenever the intrinsic dimensionality of the data manifold is small. A solution suggested by the analysis of Zhou and Belkin (2011) is to include iterated Laplacians $\boldsymbol{L}^p$ as regualrizers. It is important that our scheme applies to these more general regularizers which may not be sparse. In the case of iterated regularizers $\boldsymbol{Q} = \boldsymbol{R}^p$, our method incurs an additional quadratic dependence on $p$ and a logarithmic dependence on the generalized condition number $\kappa(\boldsymbol{R}) = ||\boldsymbol{R}||_2 ||\boldsymbol{R}^{+}||_2$ of $\boldsymbol{R}$, the following is proved in the appendix:

**Theorem 4.4.** *Given an $n \times n$ intrinsic regualarization matrix $\boldsymbol{Q} = \boldsymbol{R}^p$ where $\boldsymbol{R}$ is symmetric, diagonally dominant and has $s$ non-zero entries an approximation $\boldsymbol{A}$ to the kernel matrix $\widehat{\boldsymbol{Q}}^+$ on $\widehat{\mathcal{X}}_{\mathcal{S}}$ can be computed in expected time $\mathcal{O}\left(p\widehat{n}s \log n (\log\log n)^2 \left(\log \frac{1}{\epsilon} + p \log \kappa(\boldsymbol{R})\right) + \widehat{n}^2 n\right)$, and such that,*

$$||\boldsymbol{A} - \widehat{\boldsymbol{Q}}^+||_\infty < \epsilon + \ \textit{h.o.t.},$$

*where h.o.t. denotes terms involving order $\epsilon^2$ or higher.*

We also remark that kernels associated to the amplified resistances of von Luxburg et al. (2010) can also be efficiently approximated.

We have seen that our method will enable the construction of efficient data-dependent kernels based upon a variety of recent approaches for graph-based regularizion. We should finally mention that, when the graph is not given, forming a $k$-nearest neighbor graph, for example, can be achieved in in $\mathcal{O}(n \log n)$ (Vaidya, 1989) on low dimensional data and approximations exist for high dimensional data (Chen et al. (2009)) so there is no other computational bottleneck in this approach.

# 5 Experiments

## 5.1 Semi-supervised binary classification

We experiment on standard binary classification tasks. The first of our experiments compare the efficient semi-supervised kernels with LapSVM and a Gaussian RBF kernel SVM on the 'letter' data set from the UCI repository (Frank and Asuncion, 2010) and the 'MNIST digits' data (Lecun and Cortes). We build $k$-NN graphs with $k = 5$ and $0-1$ weights and form powers of the normalized Laplacian $\boldsymbol{L}^p_{\text{norm}}$ (with a small ridge term) as the basic intrinsic regularizer matrix $\boldsymbol{Q}$. This was mixed with an ambient Gaussian RBF kernel $K_\sigma$, with bandwidth $\sigma$, as in (4) to form the kernel for LapSVM. The subsample was chosen at random, except for a strong bias to the labeled data[5]. In our experiments we use the preconditioned conjugate gradient solver, with the preconditioner of Koutis (2011) which uses a combination of combinatorial preconditioners and multigrid methods[6], to solve the linear systems required to obtain $\widehat{\boldsymbol{Q}}$. We then form the efficient data-dependent kernel as in (6). Model selection was performed using 5-fold cross validation over a grid of values for the exponent $p$, the level of intrinsic regularization $\eta$ and the bandwidth $\sigma$ of the Gaussian kernel ($\sigma$ could alternatively be chosen using a common heuristic). The subsample is formed using $\widehat{n} = 250$ points of the labeled and unlabeled data chosen uniformly at random. All results are averaged over 50 trials. In Figure 2 we give learning curves for the three methods: the $x$-axis is the size of the labelled set. The efficient kernel recovers the performance of the full LapSVM.

In the second set of experiments, Figure 3, our set up is as above but we consider larger datasets, the full 'MNIST digits' data, on which implementing the full LapSVM is infeasible. We consider the '4 vs 9' and '3 vs 8' tasks on 12'000 labelled and unlabelled data points and the 'Odd vs Even' task on 64'000 data points (on which results are averaged over 25 trials). We consider small subsamples of size $\widehat{n} = 250$ and $\widehat{n} = 500$. We compare to the Gaussian RBF kernel and a simplistic implementation of "budget" LapSVM building a graph on the reduced subsample $\widehat{\mathcal{X}}_{\mathcal{S}}$ only, as a sanity check to ensure the method outperforms this benchmark; the point here is that in practice one would work under computational budget constraints, and the natural choice would be between discarding most of the data and implementing the full LapSVM on the reduced sample or exploiting all data with the efficient kernel measuring functions at the subsample, since they have (roughly) the same complexity in the subsample size. The efficient LapSVM substantially outperforms both the "budget" LapSVM approach and the Gaussian RBF SVM, learning much faster with, in particular, a very small labelled sample. In particular a significant advantage can be gained from the efficient method's ability to exploit all 64'000 unlabelled points in the 'Odd vs Even' task.

[5] It seems important to ensure that the labeled data does not exclusively contain points from the subsample – essentially so that cross validation is performed over some points not the domain of the intrinsic kernel, so that the algorithm does not learn the transductive problem, but a precise ratio seems unimportant.

[6] This is a practical solver using combinatorial preconditioners, though not the implementation achieving nearly-linear theoretical performance.

## 5.2 Clustering

Another application of the efficiently computable data-dependent kernel is to clustering. We consider 2 class clustering on an artificially generated 2-moons data set with 1000 data points. For a kernel $K : \mathcal{X} \times \mathcal{X} \to \mathbb{R}$ we define a metric $d$ on $\mathcal{X}$ via $d(x, x') = ||K(x, \cdot) - K(x', \cdot)||_K = \sqrt{K(x, x) + K(x', x') - 2K(x, x')}$. We investigate $k$-means clustering ($k = 2$) comparing the full LapSVM kernel, the efficient data-dependent kernel (generated as outlined in Section 5.1, with $p = 2$) and Euclidean distance. The efficient kernel uses a subsample $\widehat{\mathcal{X}}_{\mathcal{S}}$ of size $\widehat{n} = 40$ to measure functions, whereas the full LapSVM kernel uses all 1000 data points. We selected the best kernels from a small grid over the parameters $\gamma$ and $\eta$. The results are displayed in Figure 4: the Euclidean distances incurred an error of 11.4%, the full LapSVM kernel achieved perfect clustering with 0% and the efficient kernel achieved 1% error. Thus using a subsample of just 4% of the data, we are able to almost recover, in nearly-linear rather than cubic time, the performance of the full LapSVM kernel.

| Dataset | sample size $n$ (labeled + unlabeled) | subsample size $\widehat{n} = \lvert\widehat{\mathcal{X}}_{\mathcal{S}}\rvert$ | test set size |
|---|---|---|---|
| letter D vs O | 1250 | 250 | 308 |
| letter O vs Q | 1250 | 250 | 286 |
| MNIST 2 vs 3 | 2000 | 250 | 405 |
| MNIST 3 vs 8 | 12'000 | 250 | 1966 |
| MNIST 4 vs 9 | 12'000 | 250 | 1782 |
| MNIST Odd vs Even | 64'000 | 500 | 500 |

Table 1: Binary classification experiments

## 5.3 Practical timing results

To validate the practical timing performance of the proposed method we consider the time taken to compute the semi-supervised kernels on the MNIST data, as detailed in Section 5.1 but using a non-normalised Laplacian and $p = 1$, $\gamma = 1$ and $\eta = 1$. We consider the computation time of the inverse of the $\widehat{n} \times \widehat{n}$ matrix $(\boldsymbol{I}_{\widehat{n}} + \eta\widehat{\boldsymbol{Q}}\widehat{\boldsymbol{K}})^{-1}$, including the computation of the matrix $\widehat{\boldsymbol{Q}}$ from $\boldsymbol{Q}$, which is the heart of the efficient kernel computation, and theoretically nearly-linear. We compare 2 methods of solving the linear systems required to compute $\widehat{\boldsymbol{Q}}$: the preconditioned conjugate gradient solver, with the combinatorial preconditioner of Koutis (2011) used in the experiments; and the Matlab "backslash" operator. We compare these results to the computation of the inverse of the (non-sparse) $n \times n$ matrix $(\boldsymbol{I}_n + \eta\boldsymbol{Q}\boldsymbol{K})^{-1}$ which is the computational bottleneck of the standard semi-supervised kernel construction, and is cubic in complexity.

Results are shown in Figure 5: in practice the method is extremely fast, the efficiently computable kernels can be computed on 64'000 MNIST data points in 3 minutes (and the computation remains feasible on much larger data still). The preconditioned conjugate gradient method achieves approximately linear complexity in our experiments. The backslash method is also very fast on small data sizes (presumably due to the vectorization of the Matlab implementation) but appears to be growing super-linearly on this data set. As expected, the computation time of the standard semi-supervised kernel construction becomes infeasible for just a few thousand data points.

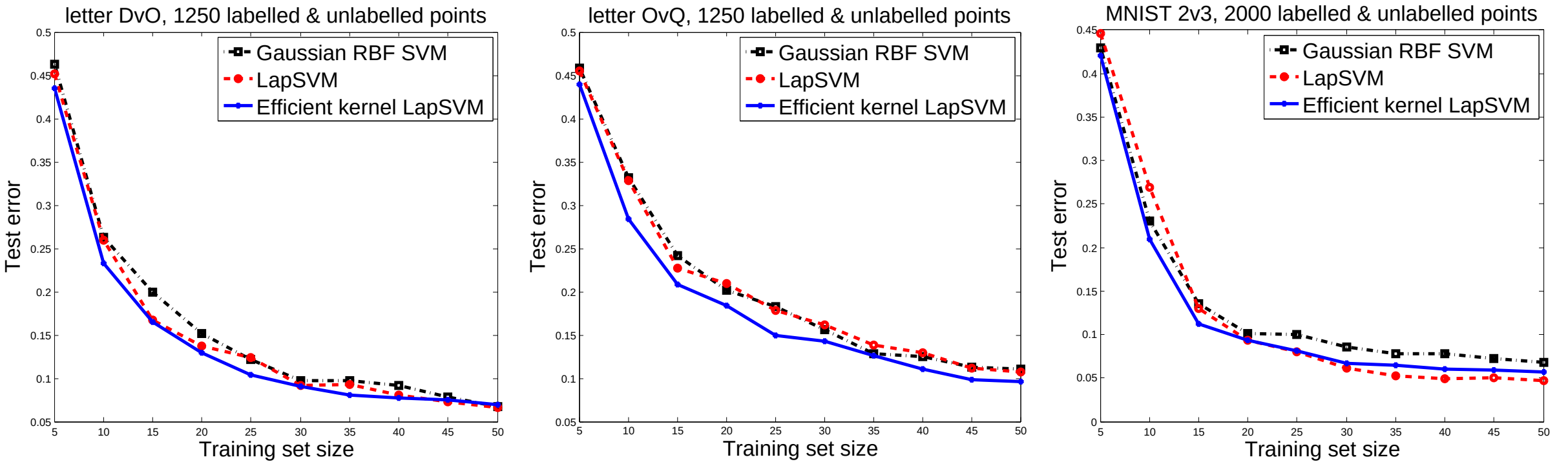

(a) letter D vs O (b) letter O vs Q (c) MNIST 2 vs 3

Figure 2: Classification: small data sets

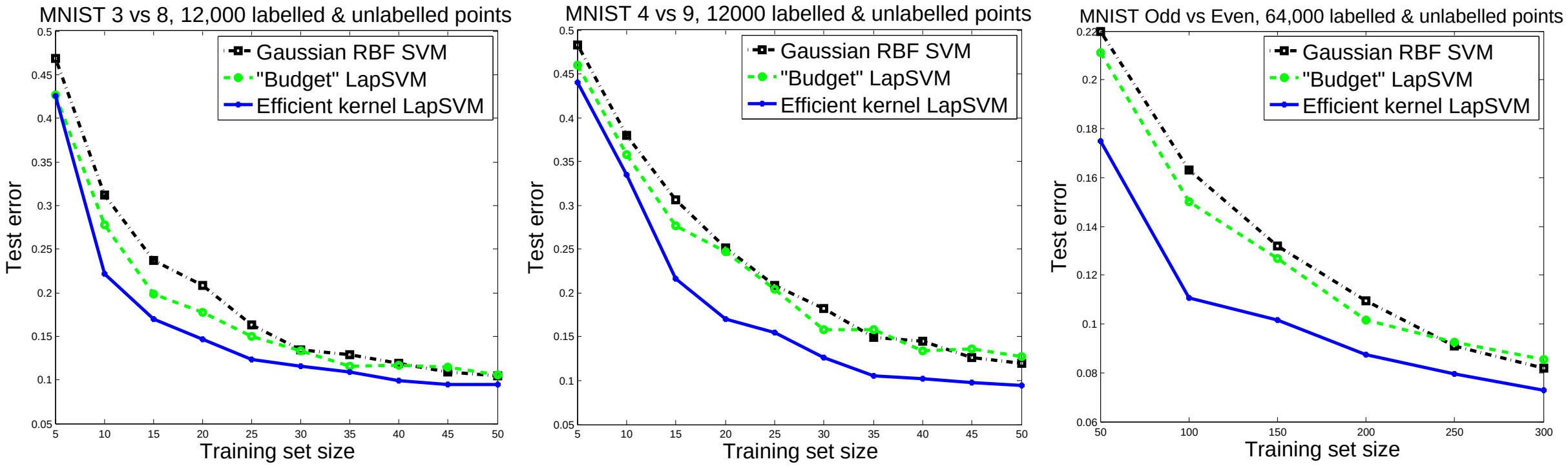

(a) MNIST 3 vs 8 (b) MNIST 4 vs 9 (c) MNIST Odd vs Even

Figure 3: Classification: large data sets

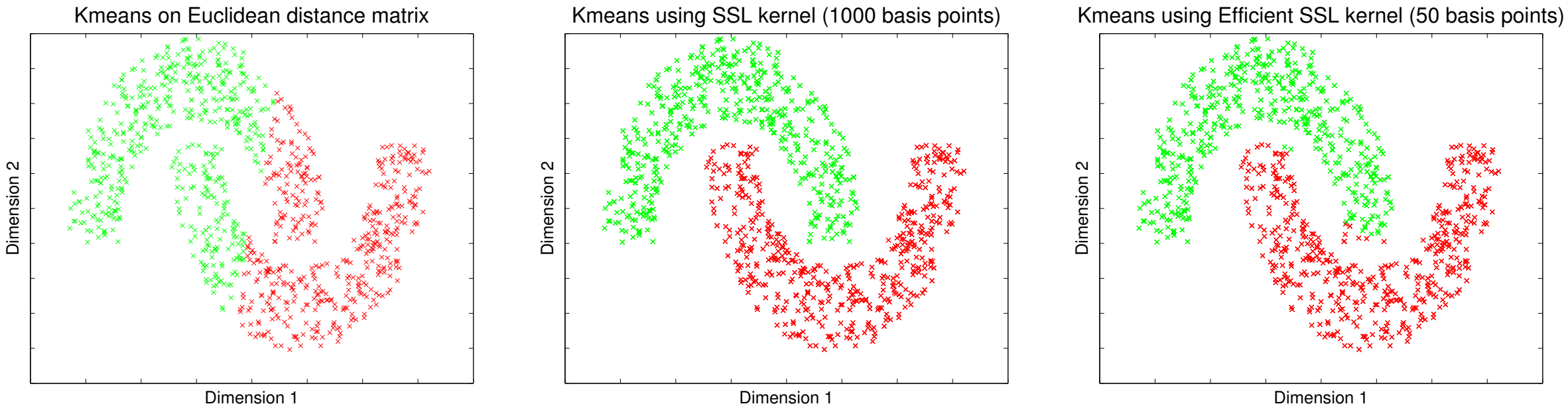

(a) Euclidean distance (b) Full LapSVM kernel (c) Efficient kernel

Figure 4: Clustering: 2-moons data, 1000 data points

# 6 Conclusions

We have presented a method for generating data-dependent kernels in nearly-linear time. The method is based on disconnecting the number of data points used to build a data-dependent regularization matrix and the number of points at which functions are measured. By measuring at fewer points and (implicitly) interpolating, our method is able to exploit huge amounts of unlabelled data in semi-supervised and unsupervised learning tasks.

Encouragingly, our experiments show that a significant advantage can be gained in semi-supervised learning from the ability to exploit a much greater quantity of unlabelled data: on large datasets of 64'000 data points the advantage gained from exploiting the large quantity of unlabelled data is clear, and much greater than the improvement demonstrated when only a small quantity of unlabelled data can be exploited. In a clustering experiment the method approximately recovers the performance of the full kernel by measuring functions at a small fraction the datapoints.

# References

M. Belkin, I. Matveeva, and P. Niyogi. Regularization and semi-supervised learning on large graphs. In *COLT*, pages 624–638, 2004.

M. Belkin, P. Niyogi, and V. Sindhwani. Manifold regularization: A geometric framework for learning from labeled and unlabeled examples. *Journal of Machine Learning Research*, 7:2399–2434, 2006.

J. Chen, H. ren Fang, and Y. Saad. Fast approximate nn graph construction for high dimensional data via recursive lanczos bisection. *Journal of Machine Learning Research*, 10:1989–2012, 2009.

R. Collobert, F. H. Sinz, J. Weston, and L. Bottou. Large scale transductive svms. *Journal of Machine Learning Research*, 7:1687–1712, 2006.

A. Frank and A. Asuncion. UCI machine learning repository, 2010. URL `http://archive.ics.uci.edu/ml`.

J. Garcke and M. Griebel. Semi-supervised learning with sparse grids. In *Proc. of the 22nd ICML Workshop on Learning with Partially Classified Training Data*, 2005.

R. A. Horn and C. R. Johnson. *Matrix Analysis*. Cambridge University Press, 1990.

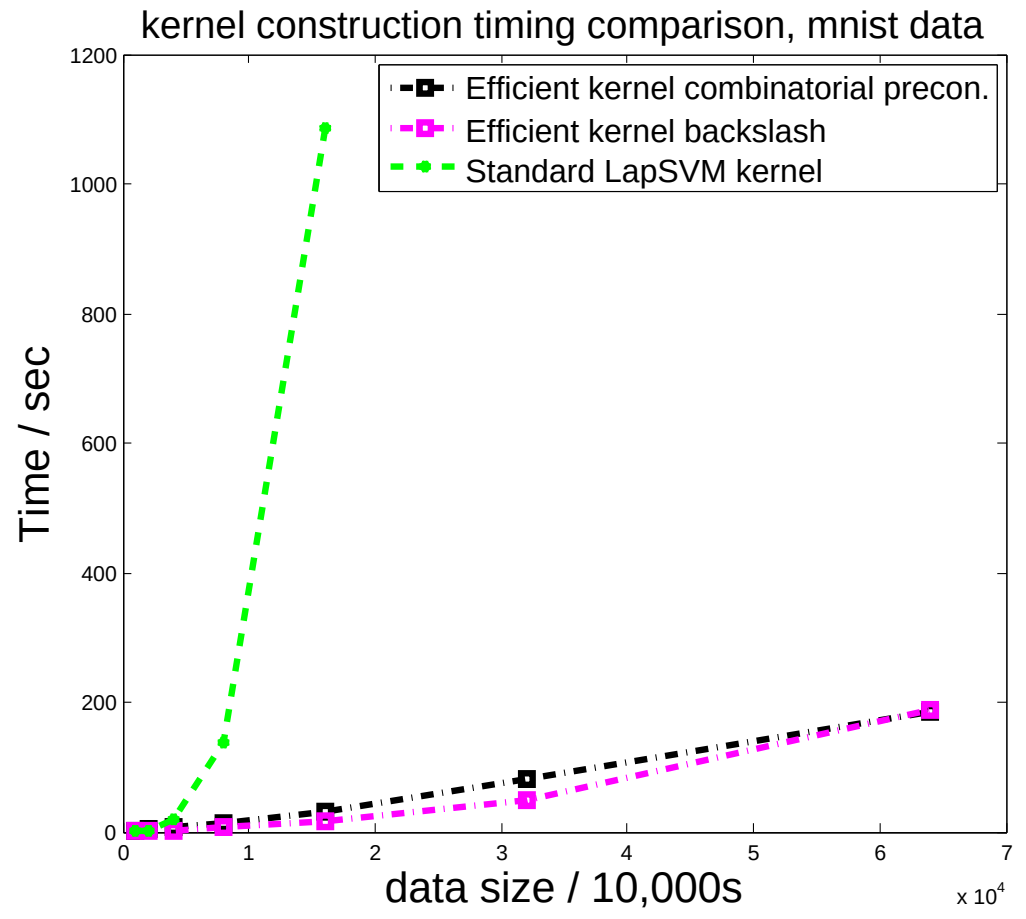

Figure 5: Computation time of LapSVM and efficient data-dependent kernels

I. Koutis. Cmg: Combinatorial multigrid solver. 2011. URL `http://www.cs.cmu.edu/~jkoutis/cmg.html`.

I. Koutis, G. L. Miller, and R. Peng. Approaching optimality for solving sdd linear systems. In *FOCS*, pages 235–244, 2010.

I. Koutis, G. L. Miller, and R. Peng. Solving sdd linear systems in time $\tilde{O}(m \log n \log(1/\epsilon))$. *CoRR*, abs/1102.4842, 2011.

Y. Lecun and C. Cortes. The mnist database of handwritten digits. URL `http://yann.lecun.com/exdb/mnist/`.

M. Mahdaviani, N. de Freitas, B. Fraser, and F. Hamze. Fast computational methods for visually guided robots. In *ICRA*, pages 138–143, 2005.

S. Melacci and M. Belkin. Laplacian support vector machines trained in the primal. *Journal of Machine Learning Research*, 12: 1149–1184, 2011.

B. Nadler, N. Srebro, and X. Zhou. Statistical analysis of semi-supervised learning: The limit of infinite unlabelled data. In Y. Bengio, D. Schuurmans, J. Lafferty, C. K. I. Williams, and A. Culotta, editors, *Advances in Neural Information Processing Systems 22*, pages 1330–1338. 2009.

C. A. Rohde. Generalized inverses of partitioned matrices. *Journal of the society of industrial and applied mathematics*, 13(4): 1033–1035, 1965.

V. Sindhwani and S. S. Keerthi. Large scale semi-supervised linear svms. In *SIGIR*, pages 477–484, 2006.

V. Sindhwani, P. Niyogi, and M. Belkin. Beyond the point cloud: from transductive to semi-supervised learning. In *ICML*, pages 824–831, 2005.

A. J. Smola and R. I. Kondor. Kernels and regularization on graphs. In *COLT*, pages 144–158, 2003.

A. J. Smola, B. Schölkopf, and K.-R. Müller. The connection between regularization operators and support vector kernels. *Neural Netw.*, 11(4):637–649, 1998.

D. A. Spielman and S.-H. Teng. Nearly-linear time algorithms for preconditioning and solving symmetric, diagonally dominant linear systems. *CoRR*, abs/cs/0607105, 2006.

I. W. Tsang and J. T. Kwok. Large-scale sparsified manifold regularization. In *NIPS*, pages 1401–1408, 2006.

P. M. Vaidya. An o(n log n) algorithm for the all-nearest.neighbors problem. *Discrete & Computational Geometry*, 4:101–115, 1989.

U. von Luxburg, A. Radl, and M. Hein. Getting lost in space: Large sample analysis of the resistance distance. In J. Lafferty, C. K. I. Williams, J. Shawe-Taylor, R. Zemel, and A. Culotta, editors, *Advances in Neural Information Processing Systems 23*, pages 2622–2630. 2010.

X. Zhou and M. Belkin. Semi-supervised learning by higher order regularization. In *AISTATS*, 2011.

X. Zhu and J. D. Lafferty. Harmonic mixtures: combining mixture models and graph-based methods for inductive and scalable semi-supervised learning. In *ICML*, pages 1052–1059, 2005.

X. Zhu, Z. Ghahramani, and J. Lafferty. Semi-supervised learning using gaussian fields and harmonic functions. In *ICML*, pages 912–919, 2003.

# A Proofs

## A.1 Proof of Theorem3.1

*Proof.* We just need to check the reproducing property of $K = \boldsymbol{R}^{+}$ for all $v \in \mathcal{V}$ and $\boldsymbol{h} \in \mathrm{im}(\boldsymbol{R})$: $\langle \boldsymbol{h}, K(v_i, \cdot)\rangle_{\mathcal{H}} = \langle \boldsymbol{h}, \boldsymbol{R}^{+}\boldsymbol{e}_i\rangle_{\mathcal{H}} = \boldsymbol{h}^{\top}\boldsymbol{R}\boldsymbol{R}^{+}\boldsymbol{e}_i = \boldsymbol{h}^{\top}\boldsymbol{e}_i = h_i = h(v_i)$. □

## A.2 Proof of Theorem 4.3

*Proof.* Theorem 4.2 implies that in time $\mathcal{O}(\widehat{n}s \log n (\log\log n)^2 \log \frac{q\widehat{n}\eta\kappa}{\epsilon\lambda_{\min}^2} + \widehat{n}^2 n)$, we can compute an $\boldsymbol{A}$ such that,

$$||(\boldsymbol{A} + \eta\boldsymbol{K}) - (\widehat{\boldsymbol{Q}}^{-1} + \eta\boldsymbol{K})||_2 < \frac{\epsilon\lambda_{\min}^2}{\eta\widehat{n}\kappa} \tag{13}$$

which implies that (see for example Horn and Johnson, 1990, section 5.8),

$$||(\boldsymbol{A} + \eta\boldsymbol{K})^{-1} - (\widehat{\boldsymbol{Q}}^{-1} + \eta\boldsymbol{K})^{-1}||_2 < \frac{\epsilon}{\eta\widehat{n}\kappa} + h.o.t. \tag{14}$$

where h.o.t. denotes terms in $\epsilon^2$ or greater. Define $\breve{K}_{\boldsymbol{A}}(x, x') := K(x, x') + \eta\widehat{\boldsymbol{k}}_x^{\top}(\boldsymbol{A} + \eta\widehat{\boldsymbol{K}})^{-1}\widehat{\boldsymbol{k}}_{x'}$. Then since $\sup_{x\in\mathcal{X}} ||\widehat{\boldsymbol{k}}_x||^2 = \kappa\widehat{n}$ and since $\widehat{\boldsymbol{Q}}$ is positive definite,

$$\begin{aligned}
\sup_{x,x'\in\mathcal{X}} |\breve{K}_{\boldsymbol{A}}(x, x') - \breve{K}(x, x')| &= \eta \sup_{x,x'\in\mathcal{X}} |\widehat{\boldsymbol{k}}_x^{\top}\left((\boldsymbol{A} + \eta\boldsymbol{K})^{-1} - (\widehat{\boldsymbol{Q}}^{-1} + \eta\boldsymbol{K})^{-1}\right)\widehat{\boldsymbol{k}}_{x'}| \\
&\leq \eta||(\boldsymbol{A} + \eta\boldsymbol{K})^{-1} - (\widehat{\boldsymbol{Q}}^{-1} + \eta\boldsymbol{K})^{-1}||_2 \sup_{x\in\mathcal{X}} ||\widehat{\boldsymbol{k}}_x||^2 \\
&\leq \epsilon + h.o.t.
\end{aligned}$$

□

## A.3 Proof of Theroem 4.4

*Proof.* We begin by making $p\widehat{n}$ calls to the solver of Koutis et al. (2011) to iteratively solve the equations

$$\boldsymbol{R}\boldsymbol{z}_i^{(j)} = \boldsymbol{z}_i^{(j-1)}$$

for each $i$ where $\boldsymbol{x}_{s_i} \in \widehat{\mathcal{X}}_{\mathcal{S}}$ and all $1 \leq j \leq p$ and where $\boldsymbol{z}_i^{(0)} = \boldsymbol{e}_{s_i}$. This gives $\boldsymbol{z}_i^{(j)} = \boldsymbol{R}^{+}\boldsymbol{z}_i^{(j-1)} + \boldsymbol{r}_i^{(j)}$ such that

$$||\boldsymbol{r}_i^{(j)}||_{\boldsymbol{R}} \leq \epsilon||\boldsymbol{R}^{+}\boldsymbol{z}_i^{(j-1)}||_{\boldsymbol{R}}, \tag{15}$$

in total time $\mathcal{O}(p\widehat{n}s \log n (\log\log n)^2 \log \frac{1}{\epsilon})$ by Lemma 4.1. Now note that,

$$\begin{aligned}
\boldsymbol{z}_i^{(j)} &= \boldsymbol{R}^{+}\boldsymbol{z}_i^{(j-1)} + \boldsymbol{r}_i^{(j)} \\
&= \boldsymbol{R}^{+}(\boldsymbol{R}^{+}\boldsymbol{z}_i^{(j-2)} + \boldsymbol{r}_i^{(j-1)}) + \boldsymbol{r}_i^{(j)} \\
&= (\boldsymbol{R}^{+})^{j}\boldsymbol{z}_i^{(0)} + (\boldsymbol{R}^{+})^{j-1}\boldsymbol{r}_i^{(1)} + \ldots + \boldsymbol{R}^{+}\boldsymbol{r}_i^{(j-1)} + \boldsymbol{r}_i^{(j)}.
\end{aligned}$$

Thus,

$$\begin{aligned}||\boldsymbol{z}_i^{(j)} - (\boldsymbol{R}^{+})^j \boldsymbol{e}_{s_i}||_{\boldsymbol{R}} &\leq ||(\boldsymbol{R}^{+})^{j-1}\boldsymbol{r}_i^{(1)}||_{\boldsymbol{R}} + \ldots + ||\boldsymbol{R}^{+}\boldsymbol{r}_i^{(j-1)}||_{\boldsymbol{R}} + ||\boldsymbol{r}_i^{(j)}||_{\boldsymbol{R}} \\ &\leq ||\boldsymbol{R}^{+}||_2^{j-1}||\boldsymbol{r}_i^{(1)}||_{\boldsymbol{R}} + \ldots + ||\boldsymbol{R}^{+}||_2||\boldsymbol{r}_i^{(j-1)}||_{\boldsymbol{R}} + ||\boldsymbol{r}_i^{(j)}||_{\boldsymbol{R}}\end{aligned} \tag{16}$$

Now note, by repeatedly applying (15),

$$\begin{aligned}||\boldsymbol{r}_i^{(k)}||_{\boldsymbol{R}} &\leq \epsilon||\boldsymbol{R}^{+}\boldsymbol{z}_i^{(k-1)}||_{\boldsymbol{R}} \\ &\leq \epsilon||\boldsymbol{R}^{+}||_2||\boldsymbol{z}_i^{(k-1)}||_{\boldsymbol{R}} \\ &\leq \epsilon||\boldsymbol{R}^{+}||_2\left(||\boldsymbol{R}^{+}\boldsymbol{z}_i^{(k-2)}||_{\boldsymbol{R}} + ||\boldsymbol{r}_i^{(k-1)}||_{\boldsymbol{R}}\right) \\ &\leq \epsilon||\boldsymbol{R}^{+}||_2\left(||\boldsymbol{R}^{+}||_2||\boldsymbol{z}_i^{(k-2)}||_{\boldsymbol{R}} + \epsilon||\boldsymbol{R}^{+}\boldsymbol{z}_i^{(k-2)}||_{\boldsymbol{R}}\right), \\ &\leq \epsilon||\boldsymbol{R}^{+}||_2^2||\boldsymbol{z}_i^{(k-2)}||_{\boldsymbol{R}} + h.o.t. \\ &\vdots \\ &\leq \epsilon||\boldsymbol{R}^{+}||_2^{k-1}||\boldsymbol{z}_i^{(1)}||_{\boldsymbol{R}} + h.o.t. \\ &\leq \epsilon||\boldsymbol{R}^{+}||_2^{k-1}||\boldsymbol{R}^{+}\boldsymbol{e}_{s_i}||_{\boldsymbol{R}} + h.o.t. \\ &\leq \epsilon||\boldsymbol{R}^{+}||_2^{k-\frac{1}{2}} + h.o.t.,\end{aligned} \tag{17}$$

and so plugging this into (16) gives,

$$\begin{aligned}||\boldsymbol{z}_i^{(j)} - (\boldsymbol{R}^{+})^j \boldsymbol{e}_{s_i}||_{\boldsymbol{R}} &\leq j\epsilon||\boldsymbol{R}^{+}||_2^{(j-\frac{1}{2})} + h.o.t. \\ ||\boldsymbol{z}_i^{(j)} - (\boldsymbol{R}^{+})^j \boldsymbol{e}_{s_i}||_{\boldsymbol{R}^j} &\leq j\epsilon||\boldsymbol{R}||_2^{(j-1)/2}||\boldsymbol{R}^{+}||_2^{(j-\frac{1}{2})} + h.o.t. \\ ||\boldsymbol{z}_i^{(p)} - \boldsymbol{Q}^{+}\boldsymbol{e}_{s_i}||_{\boldsymbol{Q}} &\leq p\epsilon||\boldsymbol{R}||_2^{(p-1)/2}||\boldsymbol{R}^{+}||_2^{(p-\frac{1}{2})} + h.o.t.\end{aligned}$$

Now let $\boldsymbol{Z} := \begin{pmatrix} \boldsymbol{z}_1^{(p)} & \ldots & \boldsymbol{z}_{\widehat{n}}^{(p)} \end{pmatrix}$ and

$$\boldsymbol{A} := \boldsymbol{Z}^{\top}\boldsymbol{Q}\boldsymbol{Z} = \boldsymbol{Z}^{\top}\boldsymbol{R}^p\boldsymbol{Z},$$

and note that $\boldsymbol{A}$ can be computed with $\mathcal{O}(ps\widehat{n} + \widehat{n}^2 n)$ operations since $\boldsymbol{R}$ has $s$ non-zero entries. Now note that,

$$\begin{aligned}|\widehat{Q}_{ij}^{+} - A_{ij}| &= |Q_{s_i s_j}^{+} - A_{ij}| \\ &= |\boldsymbol{e}_{s_i}^{\top}\boldsymbol{Q}^{+}\boldsymbol{e}_{s_j} - \boldsymbol{z}_i^{(p)\top}\boldsymbol{Q}\boldsymbol{z}_j^{(p)}| \\ &= |(\boldsymbol{Q}^{+}\boldsymbol{e}_{s_i} - \boldsymbol{z}_i^{(p)})^{\top}\boldsymbol{Q}\boldsymbol{Q}^{+}\boldsymbol{e}_{s_j} + (\boldsymbol{Q}^{+}\boldsymbol{e}_{s_j} - \boldsymbol{z}_j^{(p)})^{\top}\boldsymbol{Q}\boldsymbol{Q}^{+}\boldsymbol{e}_{s_i} - (\boldsymbol{Q}^{+}\boldsymbol{e}_{s_i} - \boldsymbol{z}_i^{(p)})^{\top}\boldsymbol{Q}(\boldsymbol{Q}^{+}\boldsymbol{e}_{s_j} - \boldsymbol{z}_j^{(p)})| \\ &\leq ||\boldsymbol{Q}^{+}\boldsymbol{e}_{s_i} - \boldsymbol{z}_i^{(p)}||_{\boldsymbol{Q}}\sqrt{\boldsymbol{e}_{s_j}{}^{\top}\boldsymbol{Q}^{+}\boldsymbol{e}_{s_j}} + ||\boldsymbol{Q}^{+}\boldsymbol{e}_{s_j} - \boldsymbol{z}_j^{(p)}||_{\boldsymbol{Q}}\sqrt{\boldsymbol{e}_{s_i}{}^{\top}\boldsymbol{Q}^{+}\boldsymbol{e}_{s_i}} \\ &\quad + ||\boldsymbol{Q}^{+}\boldsymbol{e}_{s_i} - \boldsymbol{z}_i^{(p)}||_{\boldsymbol{Q}}||\boldsymbol{Q}^{+}\boldsymbol{e}_{s_j} - \boldsymbol{z}_j^{(p)}||_{\boldsymbol{Q}} \\ &\leq 2p\epsilon||\boldsymbol{R}||_2^{(p-1)/2}||\boldsymbol{R}^{+}||_2^{\frac{3p-1}{2}} + h.o.t.,\end{aligned}$$

which, after setting $\epsilon'$ such that $\epsilon = 2p\epsilon'||\boldsymbol{R}||_2^{(p-1)/2}||\boldsymbol{R}^+||_2^{\frac{3p-1}{2}}$, we have that in time complexity,

$$\mathcal{O}(p\widehat{n}s\log n(\log\log n)^2\log\frac{1}{\epsilon'} + p\widehat{n}^2 s)$$
$$= \mathcal{O}(p\widehat{n}s\log n(\log\log n)^2(\log p + p\log||\boldsymbol{R}||_2||\boldsymbol{R}^+||_2 + \log\frac{1}{\epsilon}) + \widehat{n}^2 n)$$

the guarantee,

$$|\widehat{Q}^+_{ij} - A_{ij}| \leq \epsilon + h.o.t.$$

□